\documentclass[10pt,twocolumn,letterpaper]{article}

\usepackage[pagenumbers]{iccv} %
\usepackage[accsupp]{axessibility}  %

\definecolor{iccvblue}{rgb}{0.21,0.49,0.74}
\usepackage[pagebackref,breaklinks,colorlinks,allcolors=iccvblue]{hyperref}

\def\paperID{00002} %
\def\confName{ICCV}
\def\confYear{2025}

\title{RealTalk: Realistic Emotion-Aware Lifelike Talking-Head Synthesis}

\author{Wenqing Wang and Yun Fu\\
Northeastern University\\
360 Huntington Ave, Boston, MA 02115\\
{\tt\small wang.wenqin@northeastern.edu, yunfu@ece.neu.edu}
}

\begin{document}

\twocolumn[{
\renewcommand\twocolumn[1][]{#1}
\maketitle
\centering
\vspace*{-8mm   }
\includegraphics[width=0.53\textwidth]{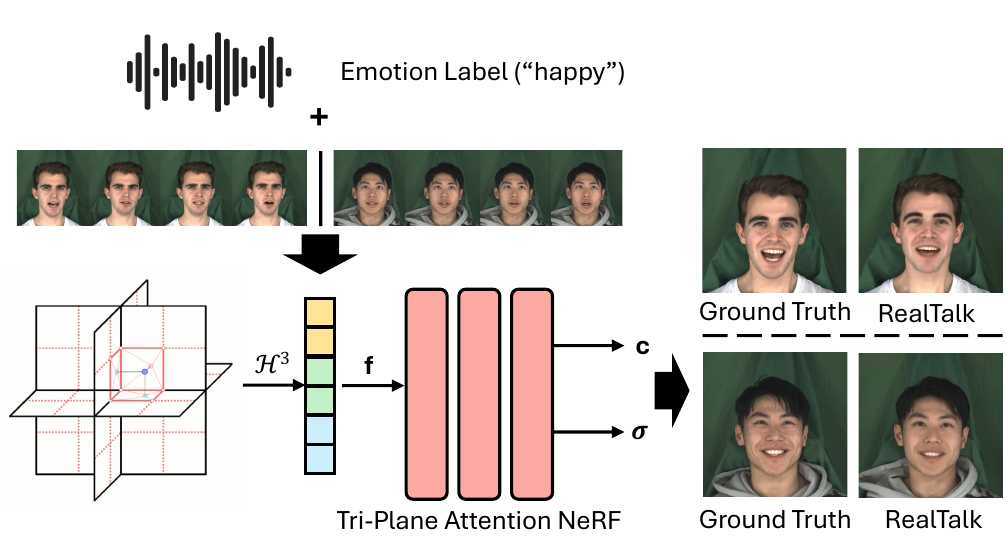}
\captionof{figure}{RealTalk generates realistic emotional talking-head videos from the input audio and emotion label.}
\label{fig:teaser}
 \vspace*{0.26cm}
}]

\maketitle

\begin{abstract}
Emotion is a critical component of artificial social intelligence. However, while current methods excel in lip synchronization and image quality, they often fail to generate accurate and controllable emotional expressions while preserving the subject’s identity. To address this challenge, we introduce RealTalk, a novel framework for synthesizing emotional talking heads with high emotion accuracy, enhanced emotion controllability, and robust identity preservation. RealTalk employs a variational autoencoder (VAE) to generate 3D facial landmarks from driving audio, which are concatenated with emotion-label embeddings using a ResNet-based landmark deformation model (LDM) to produce emotional landmarks. These landmarks and facial blendshape coefficients jointly condition a novel tri-plane attention Neural Radiance Field (NeRF) to synthesize highly realistic emotional talking heads. Extensive experiments demonstrate that RealTalk outperforms existing methods in emotion accuracy, controllability, and identity preservation, advancing the development of socially-intelligent AI systems.
\end{abstract}

\section{Introduction}
\label{sec:intro}

Generating audio-driven talking heads is a valuable and important technology for a wide range of artificial social intelligence applications, including digital avatars and virtual reality. Throughout the years, numerous methods have been introduced to synthesize talking heads from audio inputs \cite{aenerf,facetalk, NeRFFaceSpeech, geneface++,gaussiantalker,synctalk,talkinggaussian,ernerf}. Most existing methods focus on image quality and lip synchronization. However, achieving controllable and accurate emotion generation while preserving identity is an important part of realistic talking heads, which has been overlooked by these methods. Early approaches based on generative adversarial networks (GANs) \cite{wav2lip,dagan++,gogate2024robust} perform well in modeling synchronized lip motion. However, they often fail to preserve the subject's identity and typically focus on only partial facial regions. 

Recent advances in Neural Radiance Fields (NeRF) have demonstrated improvements in both image quality and fidelity for talking-head generation \cite{adnerf,aenerf,wav2nerf,geneface++,synctalk,ernerf}. Several notable NeRF-based approaches have improved lip synchronization while achieving high rendering quality \cite{geneface++, synctalk, ernerf}. Additionally, 3D Gaussian Splatting \cite{3dgs} has been recently adopted to synthesize talking-head videos by deforming 3D Gaussian attributes \cite{talkinggaussian,gaussiantalker}. Despite these advancements, the emotional dimension of generating realistic talking-head videos remains underexplored.

While existing methods \cite{emo_cai2024listen,emo_edtalk,emo_liang2022expressive,emo_style2talker,emo_wang2023emotional,eat, 11099460} have advanced emotional talking-head generation for artificial social intelligence, they face several challenges: 1) \textbf{Emotion accuracy}. They struggle to generate accurate emotional expressions, often producing vague or inconsistent results. 2) \textbf{Emotion controllability}. Many methods lack fine-grained control over specific emotions, relying on implicit emotional cues from audio or external expression sources. 3) \textbf{Identity preservation}. Existing methods struggle to maintain the subject’s identity when emotional expressions are introduced, leading to distortions and loss of key identity features. For example, EAT \cite{eat} generates noticeable emotional expressions but suffers from inconsistent regional emotions, facial distortions, and identity loss.

In this paper, we proposed \textit{RealTalk} (Figure~\ref{fig:teaser}), a novel emotion-aware talking-head generation framework to address these challenges. To handle the challenge of generating accurate emotions, we leverage facial landmarks as an intermediate representation of audio inputs, proposing a Landmark Deformation Model (LDM) with residual blocks and multi-head self-attention to transform neutral landmarks into accurate emotional landmarks. To enhance emotion controllability, we incorporate emotion labels as additional inputs, enabling precise control over the generated emotions. To address the challenge of identity preservation, we introduce a novel tri-plane NeRF with a landmark attention network to synthesize high-fidelity 3D representations of the subject. 

The main contributions of our work are summarized as follows:
\begin{itemize}
  \item We present a framework for generating realistic audio-driven talking heads with accurate, controllable emotions and preserved identity.

  \item We introduce a ResNet-based landmark deformation model with self-attention to produce precise and controllable emotional landmarks.

  \item We propose a novel tri-plane attention NeRF with a landmark attention network to render realistic emotional talking heads with preserved identity.

  \item Experiments show that our RealTalk outperforms previous works in generating high-quality emotional talking heads.
\end{itemize}

\section{Related Work}
\label{sec:related_work}

\paragraph{\textbf{Talking-Head Generation}} The task of generating audio-driven talking heads has attracted increasing attention over the years \cite{aenerf, laughtalk, audiosemantic_th,facetalk, NeRFFaceSpeech, geneface++, geneface,gaussiantalker,synctalk,talkinggaussian,ernerf}. Early methods primarily focused on synchronizing lip movements with the given audio using a static image or video \cite{lip1, lip2, wav2lip}. For instance, Wav2Lip \cite{wav2lip} employs a pre-trained lip-sync expert to generate accurate lip movements. However, these methods are often restricted to the lip region, resulting in outputs with reduced realism and poor generalization. Subsequent approaches expanded to full talking-head synthesis \cite{Zhou_2021_CVPR, audio2head, chen2021talkingheadgenerationaudio}. While GAN-based frameworks \cite{livespeechportraits,facial,wav2lip,dagan++,gogate2024robust, facial, dagan++} manage to achieve photorealistic results, they often struggle with inconsistent identity preservation and limited control over pose variations due to inadequate 3D geometry modeling. 

To enhance control over pose and facial expressions, several works have adopted landmark-based representations \cite{a2l_1,a2l_2,a2l_3,a2l_4,nocentini2023learninglandmarksmotionspeech,makeittalk}. Many of these methods leverage 3D Morphable Models (3DMM) \cite{3dmm} to capture facial landmark motion \cite{nocentini2023learninglandmarksmotionspeech, a2l_3}. Although landmark representations provide finer control, they can produce additional estimation errors that negatively affect overall quality. To address this, our proposed RealTalk framework leverages 3DMM-based landmarks with semantically informed blendshape coefficients \cite{emotalk} as the intermediate representations, improving both expression accuracy and controllability.

Due to the high-quality rendering performance of NeRF \cite{nerf}, many recent works leverage NeRF-based architectures in the talking-head generation \cite{adnerf, aenerf, wav2nerf, geneface++, geneface, synctalk, ernerf}. Standard NeRF implementations often have high computational costs, but recent advancements have addressed these limitations by employing more efficient NeRF representations, such as hash tables or sparse feature grids \cite{radnerf, ngpnerf, ernerf, synctalk}, which reduce memory overhead and hash collisions. Building on these developments, RealTalk integrates a tri-plane representation but introduces a novel landmark attention network, enabling more precise and efficient landmark conditioning for improved expression accuracy and social interaction. Recently, 3D Gaussian Splatting (3DGS) \cite{3dgs} has also been explored for talking-head synthesis \cite{talkinggaussian, gaussiantalker, gstalker, pointtalk}. Although these approaches offer efficient rendering, they often introduce artifacts and inconsistent results due to their point-based nature. Despite significant progress in NeRF and 3DGS-based methods, the integration of emotion generation remains underexplored.

\paragraph{\textbf{Emotion-Aware Talking-Head Generation}} Extending on the ideas in the talking-head generation, the emotion-aware talking-head generation aims to produce not only synchronized lip movements but also realistic emotions for enhanced social intelligence. Although previous methods have attempted to incorporate emotion generation into the talking heads \cite{emo_cai2024listen, emo_edtalk, emo_liang2022expressive, emo_style2talker, emo_wang2023emotional, laughtalk, eat, emoface, emotalk, 11099460}, achieving precise and controllable emotions while maintaining the subject’s identity remains challenging. Most of these methods either require an emotional expression source or emotional audio cues \cite{emo_wang2023emotional}, limiting direct control over emotion type and intensity. EAT \cite{eat} uses a transformer-based emotion deformation network to generate emotional expressions, but its outputs exhibit identity distortions and mismatched expressions across facial regions. EmoGene \cite{11099460} generates emotional expressions by using facial landmarks as the intermediate audio representation, deforming them into emotional landmarks for NeRF-based rendering. However, EmoGene’s landmark deformation model struggles to capture and preserve key emotional features, leading to inconsistent emotional landmarks. Additionally, its NeRF rendering quality is affected by insufficient conditional geometry features and inconsistent emotional landmarks. Inspired by this work, we also adopt deformed landmarks as an intermediate audio representation. In contrast to EmoGene, we introduce a ResNet-based LDM with self-attention and a tri-plane NeRF with a landmark attention network to overcome EmoGene's limitations in landmark consistency and geometry. In this work, we aim to generate emotional talking heads with accurate, controllable emotional expressions and preserved identity, establishing a foundation for socially-intelligent AI avatars.

\section{Method}
\label{sec:method}

\begin{figure}[!t]
  \centering
  \includegraphics[width=\columnwidth]{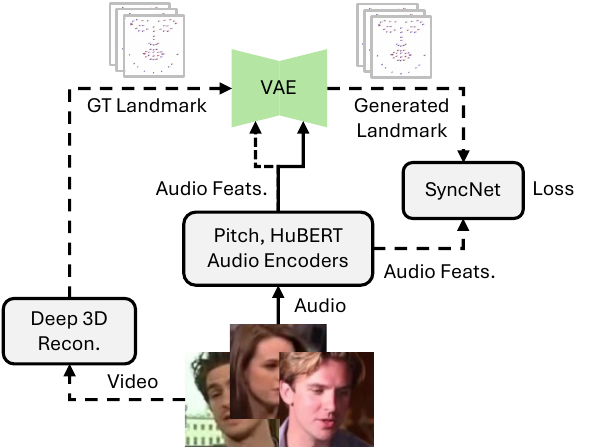}
  \caption{\textbf{The overview of audio-to-motion VAE.} Dash arrows indicate that the process is only conducted during training.}
  \label{a2m}
\end{figure}

\begin{figure*}[!t]
  \centering
  \includegraphics[width=\textwidth, trim=0 0 0 0, clip]{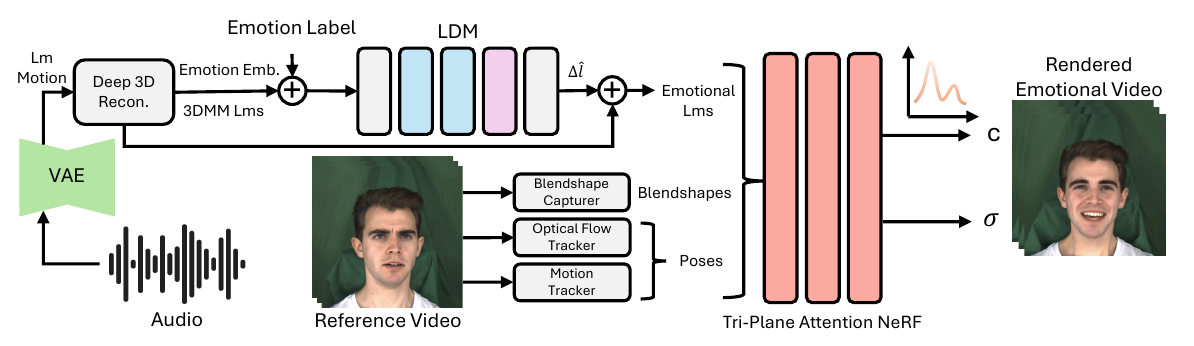}
  \caption{\textbf{The pipeline of RealTalk.} 1) The audio-to-motion VAE generates the neutral landmark from the input audio (``Lm" is the facial landmark). 2) The landmark deformation model transforms the neutral landmark into emotional landmark using the input emotion label. 3) The tri-plane attention NeRF utilizes the emotional landmarks and blendshape coefficients to generate the realistic emotional talking-head video.}
  \label{pipeline}
\end{figure*}

In this section, we introduce our proposes RealTalk framework, as shown in Figure~\ref{pipeline}. RealTalk mainly consists of three parts: 1) a VAE-based audio-to-motion model that predicts neutral facial landmark from driving audio; 2) a ResNet-based landmark deformation model that transforms the neutral landmark into emotional landmark; 3) a tri-plane attention NeRF that synthesizes a realistic emotional talking-head video from the generated emotional landmarks.

\subsection{Audio-to-Motion VAE}
We employ a variational autoencoder (VAE) to generate 3D facial landmark from the driving audio, using both audio features and ground truth (GT) facial landmark for training (Figure~\ref{a2m}).

\paragraph{\textbf{Audio Encoders}} To capture meaningful audio features, we leverage HuBERT features and pitch embeddings as our audio representations. Each feature set is passed through a dedicated audio encoder, which consists of a convolutional layer with batch normalization and GeLU activation, followed by another convolutional layer. This produces the HuBERT encoding $h$ and pitch encoding $p$. 

\paragraph{\textbf{VAE Encoder and Decoder}} The encoder and decoder of the VAE are based on convolutional layers with progressively increasing dilation factors inspired by WaveNet \cite{wavenet, geneface++}, which enables them to generate sequences with
different lengths and model temporal dependencies efficiently. At inference time, only the VAE decoder is required to generate the audio-driven landmark $\hat{l}$ using the combined audio representations $h \oplus p$ and a latent $z$:
\begin{equation}
    \hat{l} = Dec(z, h \oplus p), \quad z \sim N(0, 1).
\end{equation}

\paragraph{\textbf{Glow-based Prior}} To enhance the generative ability of the VAE and ensure a closer alignment between the latent space distribution and the real data distribution, we employ a Glow-based prior \cite{glow} as a prior of VAE.

\paragraph{\textbf{Training Process}}
We train the VAE on the audio-visual dataset VoxCeleb2 \cite{voxceleb2} by extracting audio features and corresponding 3DMM landmark using Deep 3D Face Reconstruction \cite{deep3drecon}. To guide the training of the VAE, we apply the Monte-Carlo ELBO loss \cite{elbo} along with a synchronization loss derived from a pre-trained SyncNet \cite{syncnet}. The latter ensures that the generated landmark are aligned with the audio. During the training, SyncNet takes a window of the audio clip and the generated landmark, then it calculates the synchronization score $s$ using cosine-similarity:
\begin{equation}
    s(a, \hat{l}) = \frac{a \cdot \hat{l}}{\|a\|_2 \cdot \|\hat{l}\|_2},
\end{equation}
where $a$ represents the audio features. Then, SyncNet computes the probability that the audio clip and the generated landmark are synchronized using cross-entropy loss:
\begin{equation}
    \mathcal{L}_{\text{sync}} = - \left[ y \cdot \log (s) + (1 - y) \cdot \log(1 - s) \right].
\end{equation}

The total training loss of our audio-to-motion VAE is defined as:
\begin{equation}
\mathcal{L}_{\text{VAE}} = \mathbb{E}\left[\|l - \hat{l}\|_2^2 + KL(z \mid \hat{z}) + \mathcal{L}_{\text{sync}}(a, \hat{l})\right],
\end{equation}
where $l$ denotes the ground truth landmark, $\hat{z}=Enc(l, a)$ is the latent encoding, and $z \sim N(0, 1)$. $KL$ denotes Kullback–Leibler divergence.

\begin{figure}[!t]
  \centering
  \includegraphics[width=\columnwidth]{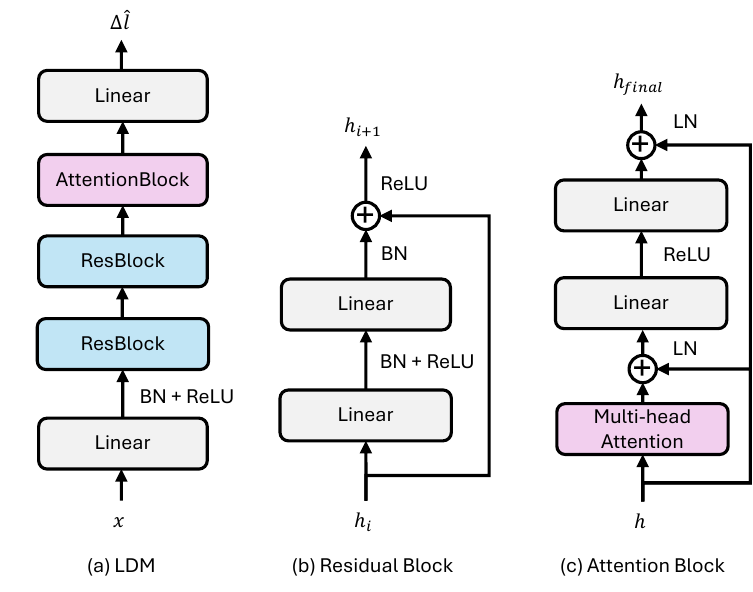}
  \caption{\textbf{The overview of landmark deformation model.}}
  \label{m2e}
\end{figure}

\begin{figure*}[!t]
  \centering
  \includegraphics[width=\linewidth]{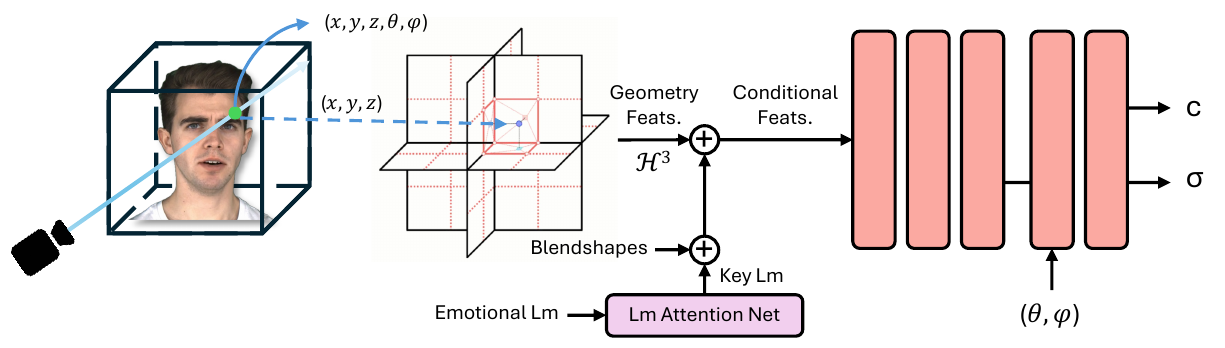}
  \caption{\textbf{The overview of tri-plane attention NeRF.}}
  \label{e2v}
\end{figure*}

\subsection{Landmark Deformation Model}
To generate consistent emotions, we leverage a ResNet-based landmark deformation model with residual attention to transform the neutral landmark into the emotion-specific landmark, using the discrete emotion labels. As shown in Figure~\ref{m2e}, the model consists of a feed-forward backbone, two residual blocks, and a residual multi-head attention block. 

\paragraph{\textbf{Feed-Forward Backbone}} Given the neutral facial landmark $\hat{l} \in \mathbb{R}^{B \times 204}$ and the emotion label $e \in \mathbb{R}^{B}$, where $B$ is the batch size, we embed $e$ into a 16-dimensional representation and concatenate them with $\hat{l}$ to produce a feature vector $x \in \mathbb{R}^{B \times 220}$. This is passed through a fully connected (FC) layer with batch normalization (BN) and ReLU activation to produce the hidden representation $h$:
\begin{equation}
    h = ReLU(BN(FC(x))).
\end{equation}

\paragraph{\textbf{Residual Blocks}} To mitigate the vanishing gradient and preserve the key features, two ResNet-style residual blocks \cite{resnet} are employed following the feed-forward backbone. As shown in Figure~\ref{m2e} (b), each residual block comprises two fully connected layers with a skip connection:
\begin{equation}
    h_{i+1} = ReLU(F(h_i) + h_i),
\end{equation}
where $F(h_i) = BN(FC(ReLU(BN(FC(h_i)))))$.

\paragraph{\textbf{Residual Attention Block}} 
To highlight key features for emotion generation, we treat each hidden representation $h$ as a single ``token" and apply a residual attention block based on the multi-head self-attention mechanism \cite{transformer}: 
\begin{equation}
    z_{\text{att}} = \text{MultiHead}(h).
\end{equation}

Our multi-head attention consists of four attention heads: 
\begin{equation}
    \text{MultiHead}(Q, K, V) = \text{Concat}(\text{head}_1, \dots, \text{head}_4) W^O,
\end{equation}
where $W$ is the output projection matrix, and each attention $\text{head}_i$ is defined as: $\text{Attention}(Q W^Q_i, K W^K_i, V W^V_i) = \text{softmax}(\frac{(Q W^Q_i)(K W^K_i)^\top}{\sqrt{d_k}}) (V W^V_i)$.

After the multi-head attention is applied, we utilize a residual skip connection and layer normalization with a 10\% dropout probability to preserve the gradients and prevent the degradation of the learned identity mappings, which provides better training efficiency:
\begin{equation}
    z = LN(Dropout(z_{\text{att}}) + h).
\end{equation}

The updated feature $z$ passes through two fully connected layers with ReLU activation, followed by a skip connection and layer normalization:
\begin{equation}
    h_{\text{final}} = LN(Dropout(FC(ReLU(FC(z)))) + h).
\end{equation}

By incorporating the residual attention block, it produces an enriched hidden representation that can capture relationships across multiple tokens for better landmark deformation.

\paragraph{\textbf{Landmark Deformation}} To produce the landmark deformation displacement $\Delta \hat{l}$ that maps the neutral landmark to the emotional landmark, a final fully connected layer is applied on $h_{\text{final}}$:  
\begin{equation}
    \Delta \hat{l} = FC(h_{\text{final}}).
\end{equation}

The resulting emotional landmark $\hat{l}_E$ is then obtained by concatenating the neutral landmark $\hat{l}$ with the emotional landmark displacement $\Delta \hat{l}$:
\begin{equation}
     \hat{l}_E = \hat{l} \oplus \delta \cdot \Delta \hat{l},
\end{equation}
where $\delta$ is a scalar controlling the deformation magnitude.

\paragraph{\textbf{Training Process}} In the training process, we train the LDM using emotional-neutral landmark pairs constructed from the MEAD dataset. The generated emotional landmark $\hat{l_E}$ is compared with the ground truth emotional landmark $l_E$ for the corresponding emotion label. The training loss for the LDM is defined as:
\begin{equation}
\mathcal{L}_{\text{LDM}} = \mathbb{E}\left[\|l_E - \hat{l_E}\|_2^2\right].
\end{equation}
\begin{figure*}[!t]
  \centering
  \includegraphics[width=\textwidth, trim=0 0 0 0, clip]{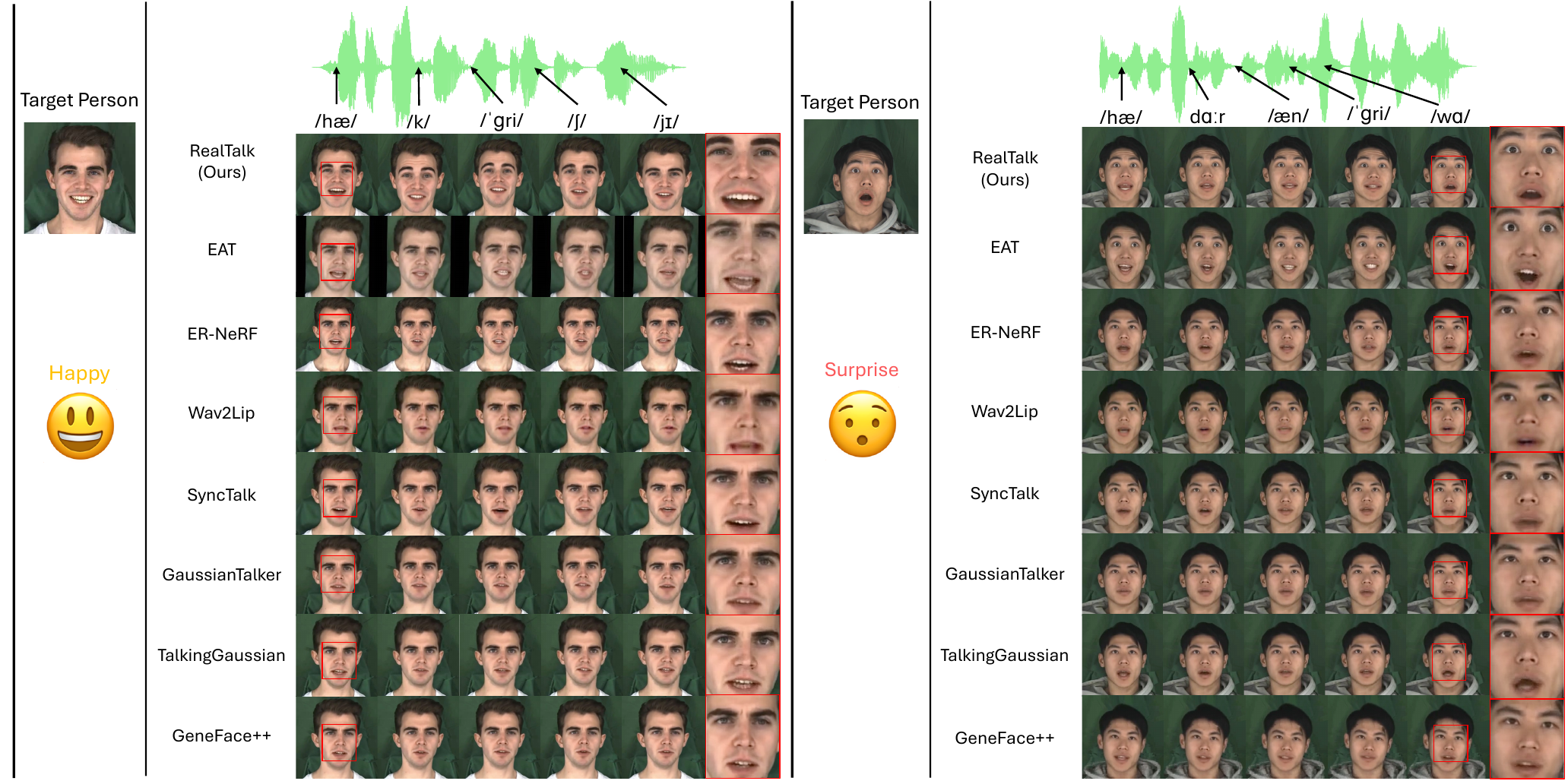}
  \caption{\textbf{Qualitative comparison of generated keyframe results.} Results of \textit{happy} and \textit{surprise} are on the left and right. The top row shows the phonetic symbol of the corresponding talking-head results generated by the evaluated methods.}
  \label{hap_sur}
\end{figure*}

\begin{table*}[!t]
\centering
\resizebox{\linewidth}{!}{%
\begin{tabular}{@{}l|cccccccc@{}}
\toprule
Criteria/Method & 
\textbf{RealTalk (Ours)} & 
Wav2Lip & 
EAT & 
GeneFace++ & 
GaussianTalker & 
SyncTalk & 
TalkingGaussian & 
ER-NeRF \\
\midrule
Emotion Acc. & 
$\mathbf{3.006 \pm 0.430}$ & 
$2.831 \pm 0.655$ & 
$2.769 \pm 0.454$ & 
$3.000 \pm 0.408$ & 
$2.550 \pm 0.558$ & 
$1.906 \pm 0.178$ & 
$1.944 \pm 0.170$ & 
$1.338 \pm 0.064$ \\
Lip Sync. & 
$2.869 \pm 0.158$ & 
$2.981 \pm 0.269$ & 
$\mathbf{3.169 \pm 0.312}$ & 
$3.144 \pm 0.346$ & 
$2.169 \pm 0.256$ & 
$1.850 \pm 0.154$ & 
$1.931 \pm 0.160$ & 
$1.300 \pm 0.139$ \\
Video Realness & 
$\mathbf{3.131 \pm 0.258}$ & 
$2.838 \pm 0.252$ & 
$2.863 \pm 0.141$ & 
$3.119 \pm 0.263$ & 
$2.481 \pm 0.196$ & 
$1.963 \pm 0.103$ & 
$2.156 \pm 0.190$ & 
$1.456 \pm 0.098$ \\
Video Quality & 
$\mathbf{3.100 \pm 0.271}$ & 
$2.856 \pm 0.188$ & 
$2.981 \pm 0.113$ & 
$3.100 \pm 0.225$ & 
$2.700 \pm 0.210$ & 
$2.225 \pm 0.222$ & 
$2.381 \pm 0.125$ & 
$1.456 \pm 0.105$ \\
\midrule
Average & 
$\mathbf{3.027 \pm 0.279}$ & 
$2.876 \pm 0.341$ & 
$2.945 \pm 0.255$ & 
$3.091 \pm 0.311$ & 
$2.475 \pm 0.305$ & 
$1.986 \pm 0.164$ & 
$2.103 \pm 0.161$ & 
$1.388 \pm 0.102$ \\
\bottomrule
\end{tabular}%
}
\caption{User study results. The best results are in \textbf{bold}.}
\label{mos}
\end{table*}

\subsection{Tri-Plane Attention NeRF}
To render high-fidelity emotional talking-head videos, we adopt a tri-plane-based attention NeRF that conditions on the generated emotional landmark and facial blendshape coefficients (Figure~\ref{e2v}).

\paragraph{\textbf{Feature Pre-processing}} 1) Following \cite{synctalk}, we extract blendshape coefficients via a Mediapipe-based capture module \cite{mediapipe}, focusing on core coefficients that relate closely to facial structure and expressions. 2) We employ an optical flow estimation model \cite{opticalflow} to track facial keypoints $K$, which are concatenated with the rotation $R$ and translation $T$ to produce the head pose. 3) To ensure a refined landmark representation, we introduce a landmark attention network consisting of several convolutional layers that learn attention weights to emphasize key landmark regions.

\paragraph{\textbf{Tri-Plane Attention NeRF}} To mitigate hash collisions in NeRF-based systems, we factorize the 3D spatial feature volume into three 2D hash grids \cite{ernerf}. For coordinates $\mathbf{x} = (x,y,z) \in \mathbb{R}^{XYZ}$, each 2D hash encoder projects the 3D point onto a plane:
\begin{equation}
    \mathcal{H}^{AB} : (a, b) \rightarrow f^{AB}_{ab},
\end{equation}
where $f^{AB}_{ab} \in \mathbb{R}^{LD}$ is the plane-level geometry feature for the projected coordinate $(a, b)$, with $L$ levels and $D$ dimensions, and $\mathcal{H}^{AB}$ is the hash encoder for plane $\mathbb{R}^{AB}$. To get the overall geometry feature $f_x = \mathbb{R}^{3 \times LD}$, we concatenate the encoded output from the three planes as:
\begin{equation}
    f_x = \mathcal{H}^{XY}(x,y) \oplus \mathcal{H}^{YZ}(y,z) \oplus \mathcal{H}^{XZ}(x,z).
\end{equation}

The input to the attention NeRF consists of $f_x$, the view direction $d$, the conditional features of the emotional landmark $\hat{l}_E$, and the blendshape coefficients $b$. The implicit function of the tri-plane attention NeRF is defined as:
\begin{equation}
    \mathcal{F}^{\mathcal{H}} \colon (x, d, \hat{l}_E, b; \mathcal{H}^3) \rightarrow (c, \sigma),
\end{equation}
where $\mathcal{H}^3 \colon x \rightarrow f_x$ is the integration of the tri-plane hash encoders, $c$ is the RGB color, and $\sigma$ is the volumn density.

The RGB color $\hat{C}$ of each pixel is rendered by aggregating along the ray $r(t)=o+t \cdot d$ from the camera origin $o$, following the differentiable volume rendering equation \cite{nerf}:
\begin{equation}
\hat{C}(r, \hat{l}_E,b) = \int_{t_n}^{t_f} \sigma(r(t), \hat{l}_E, b) \cdot c(r(t), d, \hat{l}_E, b) \cdot T(t) \, dt.
\end{equation}
The value of $\hat{C}$ is conditioned on $\hat{l}_E$ and $b$, making it adaptable to diverse emotional expressions and facial identities.

\paragraph{\textbf{Training Process}} To train the tri-plane attention NeRF, we first extract the landmark and blendshape coefficients from the video frames. Then, we sample random patches $P$ from the images and optimize a two-stage coarse-to-fine training with an MSE and LPIPS loss weighted by $\lambda_{\text{LPIPS}}$:
\begin{equation}
    \mathcal{L}_{\text{total}} = \sum_r \|C(r) - \hat{C}\|_2 + \lambda_{\text{LPIPS}} \cdot \mathcal{L}_{\text{LPIPS}}(\hat{P}, P).
\end{equation}

\section{Experiments}
\label{sec:experiments}

\begin{table*}[!t]
\centering
\resizebox{\linewidth}{!}{%
\begin{tabular}{@{}l|ccccc@{}}
\toprule
Method/Score & SSIM ↑ & PSNR ↑ & LPIPS ↓ & \(\text{Score}_{\text{emotion}} \) (\%) ↑ & M/F-LMD ↓ \\
\midrule
EAT~\cite{eat} 
    & 0.442 (-39.8\%) 
    & 13.199 (-33.3\%) 
    & 0.610 (-131.0\%) 
    &  \textcolor{red}{18.360} (+3.4\%) 
    & 9.831/26.989 (-65.9\%/-179.6\%) \\
ER-NeRF~\cite{ernerf} 
    & 0.605 (-17.5\%) 
    & 15.958 (-19.3\%) 
    & 0.515 (-95.1\%) 
    & 12.270 (-30.9\%) 
    & 6.779/19.686 (-14.4\%/-103.9\%) \\
Wav2Lip~\cite{wav2lip} 
    & 0.648 (-11.7\%) 
    & 16.369 (-17.3\%) 
    & 0.471 (-78.4\%) 
    & 12.427 (-30.0\%) 
    & \textcolor{blue}{6.610}/19.416 (-11.6\%/-101.2\%) \\
SyncTalk~\cite{synctalk} 
    & 0.656 (-10.6\%) 
    & 16.498 (-16.6\%) 
    & 0.435 (-64.6\%) 
    & 13.313 (-25.0\%) 
    & 6.705/18.681 (-13.2\%/-93.5\%) \\
GaussianTalker~\cite{gaussiantalker} 
    & 0.658 (-10.3\%) 
    & 16.290 (-17.7\%) 
    & 0.448 (-69.7\%) 
    & 12.737 (-28.2\%) 
    & 6.675/18.595 (-12.7\%/-92.7\%) \\
TalkingGaussian~\cite{talkinggaussian} 
    & 0.666 (-9.2\%) 
    & 16.761 (-15.3\%) 
    & \textcolor{blue}{0.404} (-53.0\%) 
    & 12.730 (-28.3\%) 
    & 6.632/\textcolor{blue}{16.879} (-11.9\%/-74.9\%) \\
GeneFace++~\cite{geneface++} 
    & \textcolor{blue}{0.668} (-8.9\%) 
    & \textcolor{blue}{17.182} (-13.2\%) 
    & 0.453 (-71.6\%) 
    & 11.838 (-33.3\%) 
    & 7.122/16.934 (-20.2\%/-75.4\%) \\ \midrule
\textbf{RealTalk (Ours)} 
    & \textcolor{red}{0.734} 
    & \textcolor{red}{19.780} 
    & \textcolor{red}{0.264}   
    & \textcolor{blue}{17.751}  
    & \textcolor{red}{5.925}/\textcolor{red}{9.653}  \\ \midrule
GT 
    & 1.000  
    & $\infty$  
    & 0.000  
    & 28.966 
    & 0.000/0.000  \\
\bottomrule
\end{tabular}
}
\caption{Quantitative evaluation of the compared methods. ``↑": higher is better. ``↓": lower is better. The best results are in \textcolor{red}{red}. The second best results are in \textcolor{blue}{blue}. Each cell shows the percentage difference relative to our method RealTalk.}
\label{table_quant}
\end{table*}

\paragraph{\textbf{Datasets}} We train the audio-to-motion VAE on the VoxCeleb2 dataset \cite{voxceleb2}, which comprises over one million utterances from 6,112 unique speakers. This extensive coverage enables the model to learn a generalized audio-to-motion mapping. For the landmark deformation model, we employ the MEAD dataset \cite{mead}, containing labeled emotional talking videos of 60 subjects, each exhibiting 8 distinct emotions (angry, disgust, contempt, fear, happy, neutral, sad, surprise). We partition MEAD into training and testing sets on a per-identity basis. To train the tri-plane attention NeRF, we use the MEAD dataset along with videos from \cite{adnerf}. Each training video has a duration of 3 to 6 minutes at a resolution of 512x512 resolution and 25 FPS.

\paragraph{\textbf{Compared Baselines}} We compare our RealTalk with 7 baselines: 1) a GAN-based method Wav2Lip \cite{wav2lip}; 2) a transformer-based method EAT \cite{eat}; 3) three NeRF-based methods (GeneFace++ \cite{geneface++}, SyncTalk \cite{synctalk}, ER-NeRF \cite{ernerf}); 4) two 3D Gaussian Splatting-based methods (GaussianTalker \cite{gaussiantalker}, TalkingGaussian \cite{talkinggaussian}).

\paragraph{\textbf{Implementation Details}} RealTalk is trained on a single NVIDIA RTX A6000 GPU. Convergence for the VAE and the landmark deformation model typically occurs around 40K iterations, which takes around 14 hours. The training time of the tri-plane attention NeRF is approximately 2 hours. We set $\lambda_{\text{LPIPS}}=0.1$.

\subsection{Quantitative Evaluation}

\paragraph{\textbf{Evaluation Metrics}}
We adopt the SSIM and PSNR to measure the image quality and fidelity of the generated videos. We use LPIPS to measure the perceptual image similarity of the videos. To evaluate the emotion accuracy score, we calculate the average predicted probability of the target emotion across all video frames using \textit{DeepFace}. We employ the landmark distance of mouth (M-LMD) and face (F-LMD) to measure audio-lip synchronization and expression accuracy.

\paragraph{\textbf{Evaluation Results}}
The results are presented in Table~\ref{table_quant}. We have the following observations: 1) RealTalk achieves higher image quality and fidelity, outperforming other methods in both SSIM and PSNR. Specifically, while GeneFace++ attains the second-best scores, RealTalk surpasses it by 8.9\% in SSIM and 13.2\% in PSNR. 2) RealTalk demonstrates higher perceptual image similarity, exceeding the second-best method, TalkingGaussian, by 53.0\% in LPIPS. 3) RealTalk achieves higher accuracy in emotional and expression generation, outperforming other methods in both M-LMD and F-LMD. Furthermore, it surpasses the second-best M-LMD and F-LMD by 11.6\% and 74.9\% respectively. RealTalk performs slightly worse than EAT in emotion score by 3.4\%. This may be due to its focus on preserving expression identity and fidelity, which may reduce the intensity of the generated emotions. However, RealTalk still achieves the second-best emotion score among the compared methods, surpassing the third-best method, SyncTalk, by 25.0\%.

\begin{figure*}[!t]
  \centering
  \includegraphics[width=\linewidth]{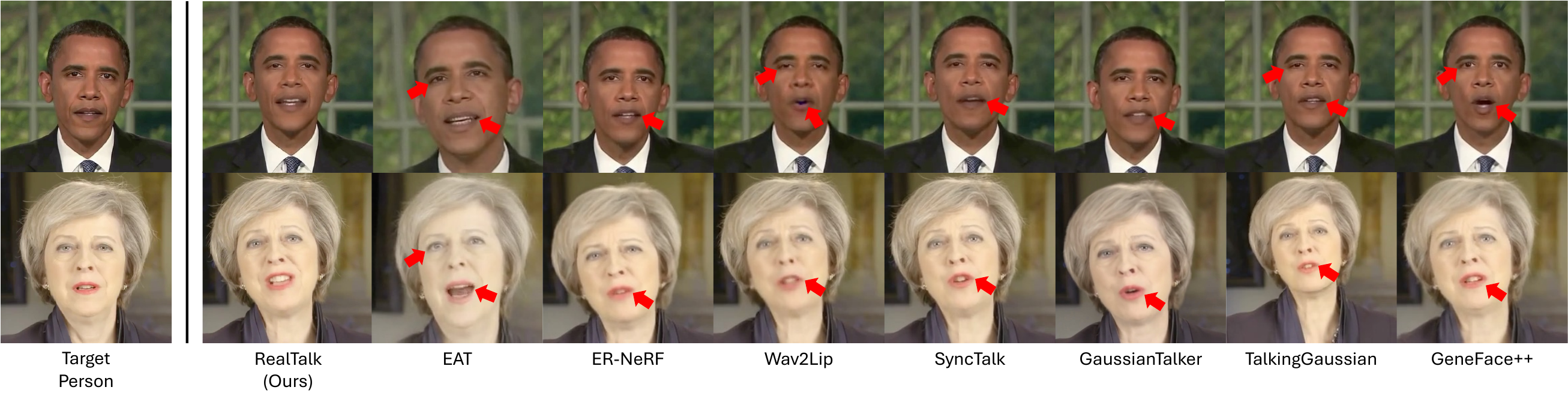}
  \caption{\textbf{Qualitative comparison of out-of-domain video generation for ``happy" emotions.}}
  \label{qua_ood}
\end{figure*}

\subsection{Qualitative Evaluation}
In Figure~\ref{hap_sur}, we compare the qualitative results of the methods by examining the generated keyframes. We observe that RealTalk stands out by producing high-resolution facial expressions that preserve the target person's identity and accurately reflect the specified emotion labels, like raised lip corners and relaxed eyes for ``happy" and open mouth with raised eyebrows for ``surprise." Conversely, EAT struggles with consistent facial features, producing unnatural expressions (e.g., lifted lips but frowned brows for ``happy") and distorted faces, causing identity loss. NeRF-based (GeneFace++, SyncTalk, ER-NeRF) and 3D Gaussian Splatting methods (GaussianTalker, TalkingGaussian) offer good reconstruction but poor emotional expression. Wav2Lip ensures lip sync but lacks emotion and produces blurry mouth results.

In addition, we show the qualitative results of the methods on the out-of-domain dataset in Figure~\ref{qua_ood}. RealTalk accurately generates ``happy" features like raised lip corners across out-of-domain datasets, while others suffer from identity loss (e.g., EAT), blurry mouths (e.g., Wav2Lip), incorrect lip motion (e.g., GaussianTalker), or exaggerated eyes (e.g., Geneface++), highlighting RealTalk’s effectiveness in producing accurate and identity-preserving emotional expressions.

\paragraph{\textbf{User Study}} We conducted a user study with 20 participants to evaluate the perceived qualities of the generated videos. For each of the 8 methods, we generated 3 video clips for each of the 8 motion categories, resulting in a total of 192 videos. We employ the Mean Opinion Score (MOS) rating for evaluation, which ranges from 1 (Bad) to 5 (Excellent). The participants are instructed to rate each video based on 4 criteria: 1) emotional accuracy; 2) lip synchronization; 3) video realness; and 4) video quality.

The average scores for each method are presented in Table~\ref{mos}. We have the following observations: 1) RealTalk outperforms the other methods by achieving the highest scores in emotional accuracy, video realness, and video quality, and the overall average score. 2) RealTalk shows slightly lower perceived lip-synchronization accuracy compared to other methods, potentially due to its emotional landmark deformation process, which may affect lip movement alignment with speech when generating emotional landmarks.

\subsection{Ablation Study}
In this section, we conduct the ablation study to prove the necessity of the proposed components in RealTalk.

\begin{table}[!t]
\centering
\resizebox{\columnwidth}{!}{%
\begin{tabular}{@{}l|ccccc@{}}
\toprule
Setting/Score & SSIM ↑ & PSNR ↑ & \( \text{Score}_{\text{emotion}} \) (\%) ↑ & LPIPS ↓ & M/F-LMD ↓ \\ 
\midrule
RealTalk [full] & 
\textbf{0.734} & \textbf{19.780} & \textbf{17.751} & \textbf{0.264} & \textbf{5.925}/9.653 \\ \midrule

w/o LDM Attention & 
0.721 & 18.201 & 12.896 & 0.285 & 5.192/\textbf{8.353} \\

w/o LDM ResBlock & 
0.721 & 18.197 & 15.568 & 0.286 & 5.534/8.539 \\ \midrule

w/o NeRF Attention & 
0.711 & 17.789 & 15.587 & 0.293 & 6.018/8.782 \\

w/o Tri-Plane Encoders & 
0.723 & 18.250 & 13.558 & 0.292 & 5.368/8.476 \\
\bottomrule
\end{tabular}%
}
\caption{Ablation study of model components. The best results are in \textbf{bold}.}
\label{abla_model}
\end{table}

\paragraph{\textbf{Ablation of LDM Components}} We test two settings of the LDM to analyze the impact of its key components (Table~\ref{abla_model}): 1) without the residual attention block, which results in noticeable decreases in SSIM, PSNR, emotion score, LPIPS, and M-LMD; 2) without residual blocks, leading to a drop in performance across all metrics. These results demonstrate the importance of both the residual attention block and the residual blocks in enhancing the performance of LDM in image quality, fidelity, and emotion accuracy.

\paragraph{\textbf{Ablation of Tri-Plane Attention NeRF Components}} We evaluate two settings of the tri-plane attention NeRF (Table~\ref{abla_model}): 1) without attention. In this setting, we remove the attention mechanism from the tri-plane attention NeRF; 2) without the tri-plane encoders. In this setting, we remove the three 2D hash encoders and replace them with a unified 3D positional hash encoder. We observe that in both settings, the performance of the NeRF dropped across all metrics, which indicates that the attention mechanism and the tri-plane 2D hash encoders are crucial for strengthening the performance of the NeRF in image quality, fidelity, and emotion accuracy.

\begin{table}[!t]
\centering
\resizebox{\columnwidth}{!}{%
\begin{tabular}{@{}l|ccccc@{}}
\toprule
Setting/Score  & SSIM ↑ & PSNR ↑ & \( \text{Score}_{\text{emotion}} \) (\%) ↑ & LPIPS ↓ & M/F-LMD ↓ \\ \midrule
0.15 [full]    & \textbf{0.734} & \textbf{19.780} & \textbf{17.751} & \textbf{0.264} & 5.925 / 9.653 \\ \midrule
0              & 0.721 & 18.155 & 12.828 & 0.285 & 5.795 / 8.624 \\
0.2            & 0.721 & 18.183 & 16.558 & 0.286 & 5.238 / 8.433 \\
0.3            & 0.720 & 18.182 & 17.352 & 0.286 & \textbf{5.018} / \textbf{8.397} \\
0.4            & 0.720 & 18.135 & 16.453 & 0.287 & 4.990 / 8.490 \\
0.5            & 0.719 & 18.118 & 17.121 & 0.288 & 5.089 / 8.507 \\
1.0              & 0.716 & 18.012 & 12.554 & 0.290 & 5.246 / 8.700 \\ \bottomrule
\end{tabular}%
}
\caption{Ablation study of LDM deformation magnitudes. The best results are in \textbf{bold}.}
\label{abla_delta}
\end{table}

\paragraph{\textbf{Ablation of Landmark Deformation Magnitudes}} To evaluate the impact of the different $\delta$ (landmark deformation magnitudes) of the LDM, we ablate six settings: $\{0, 0.2, 0.3, 0.4, 0.5, 1.0\}$ (Table~\ref{abla_delta}). We notice that the $\delta$ of $0.15$ provides the best results in SSIM, PSNR, emotion score, and LPIPS. Although the $\delta$ value of $0.3$ leads to higher M/F-LMD scores, these gains are marginal compared to the overall metrics. Furthermore, all other $\delta$ values have lower performance compared to the $\delta$ value of $0.15$. Therefore, we adopt a $\delta$ value of 0.15 in our experiments.

\section{Conclusion}
\label{sec:conclusion}
This paper presents RealTalk, a novel framework for generating realistic emotional talking-head videos with controllable emotions and preserved identity. We propose a landmark deformation model with attention mechanisms to generate robust emotional landmarks, conditioning a tri-plane attention NeRF for rendering lifelike, identity-preserving talking heads. However, RealTalk is limited by predefined emotion categories, restricting its adaptability to complex emotions, and requires per-subject training, limiting scalability. Future work could focus on expanding the emotional training dataset to encompass a wider spectrum of emotions, which may improve the range of generative emotions. To improve scalability, future works could investigate alternative generative methods and refine the underlying architectural components. These advancements could further improve the versatility and efficiency of emotional talking-head generation, paving the way for more dynamic and widely applicable artificial social intelligence.

{
    \small
    \bibliographystyle{ieeenat_fullname}
    \bibliography{main}
}

\end{document}